\documentclass{article}

\usepackage{PRIMEarxiv}

\usepackage[utf8]{inputenc} 
\usepackage[T1]{fontenc}    
\usepackage{amsmath,amssymb,amsfonts}%
\usepackage{booktabs}
\usepackage{multirow} %
\usepackage{algorithm}  
\usepackage{algpseudocode}
\usepackage{graphicx}  
\usepackage{enumitem}  
\usepackage[export]{adjustbox}
\usepackage{subfigure}
\usepackage[utf8]{inputenc}
\usepackage{microtype}  
\usepackage{verbatim}  
\usepackage{pythonhighlight}  
\usepackage{wrapfig}  
\usepackage{caption}
\usepackage{appendix} 
\usepackage{float}  
\usepackage{subcaption}  
\usepackage{placeins}
\usepackage{url}
\usepackage{wrapfig}
\usepackage{booktabs}  
\usepackage{float}

\graphicspath{{figures}}  

\title{Counterfactual experience augmented off-policy reinforcement learning
}

\author{
  Sunbowen Lee\thanks{First author}, Yicheng Gong\thanks{Corresponding author} \\
  Hubei Province Key Laboratory of \\ System Science in Metallurgical Process \\
  College of Science \\
  Wuhan University of Science and Technology \\
  Wuhan, China\\
  \texttt{\{bw1863, gongyicheng\}@wust.edu.cn} \\
   \And
  Chao Deng \\
  Intelligent Automobile Engineering Research Institute \\
  School of Automobile and Traffic Engineering\\
  Wuhan University of Science and Technology \\
  Wuhan, China\\
  \texttt{woec@wust.edu.cn} \\
}

\begin{document}
\maketitle

\begin{abstract}
Reinforcement learning control algorithms face significant challenges due to out-of-distribution and inefficient exploration problems. While model-based reinforcement learning enhances the agent's reasoning and planning capabilities by constructing virtual environments, training such virtual environments can be very complex. In order to build an efficient inference model and enhance the representativeness of learning data, we propose the Counterfactual Experience Augmentation (CEA) algorithm. CEA leverages variational autoencoders to model the dynamic patterns of state transitions and introduces randomness to model non-stationarity. This approach focuses on expanding the learning data in the experience pool through counterfactual inference and performs exceptionally well in environments that follow the bisimulation assumption. Environments with bisimulation properties are usually represented by discrete observation and action spaces, we propose a sampling method based on maximum kernel density estimation entropy to extend CEA to various environments. By providing reward signals for counterfactual state transitions based on real information, CEA constructs a complete counterfactual experience to alleviate the out-of-distribution problem of the learning data, and outperforms general SOTA algorithms in environments with difference properties. Finally, we discuss the similarities, differences and properties of generated counterfactual experiences and real experiences. The code is available at \url{https://github.com/Aegis1863/CEA}.
\end{abstract}


\section{Introduction}\label{Introduction}

Advanced control strategies have attracted substantial interest, particularly with recent progress in deep learning and reinforcement learning (RL). But all learning-based algorithms face the out-of-distribution (OOD) problem. In the field of online RL, data must be collected from the environment in real time without domain knowledge, making it very difficult for RL agents to learn comprehensive policies in underrepresented state distributions.


There are many attempts to solve the OOD problem. For example, RAD \cite{RAD}, DBC \cite{dbc}, DrAC \cite{DrAC}, DrQ \cite{drq, drqv2}, SODA \cite{SODA}, etc., these methods effectively make use of representation learning and data enhancement methods. They transfer the image enhancement method in computer vision to vision-based RL, achieving a robust and efficient control effect. They are usually applied to robot control, but not for tasks with vector-based observation. Specifically, most of them enhance the robustness of convolutional parameters by rotating or adding noise to the observed images. However, for a dense vector, these methods will destroy its information. The data enhancement of dense vectors requires a more conservative and cautious approach. We noticed a related work based on vector observation augmentation, Acamda \cite{Sun2024}, but there is no code, and it is difficult to reproduce. Therefore, we provide an executable method with open-source code.

\begin{wrapfigure}{r}{0.5\textwidth}
	\centering
	\includegraphics[width=0.9\linewidth]{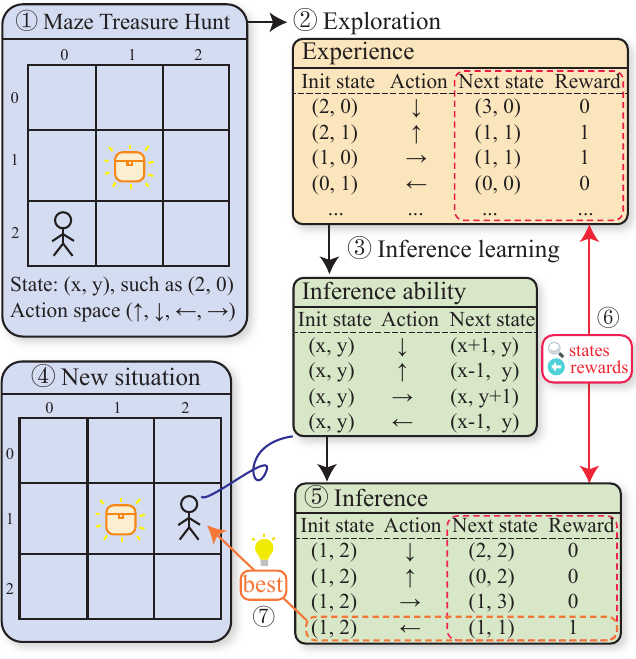}
	\caption{Counterfactual inference learning. The agent's task is to find the treasure chest. When the agent moves to a state with the same coordinate as the treasure chest, the reward increases by 1, otherwise remains 0. We marked the entire process with serial numbers in the figure. In the first and second steps, the agent randomly selects its action and records the process of position transfer. This record can be used for inference learning, with its ability referring to table "Inference ability". Once this inference ability is obtained, it can be used to infer the results of the corresponding actions in new situations like step four. The agent can compare the predicted results with historical information, as shown in step six, thereby choosing the best action without actually executing the action. The action space and state space of the environment are both discrete, which satisfies the bisimulation assumption.}\label{CEA_figure}
\end{wrapfigure}

To effectively handle dense vectors, autoencoders have been regarded as effective tools. For example, sequential encoders \cite{hafner2019learning, NEURIPS202008058bf5, YaratsZhangKostrikovAmosPineauFergus2021} are implemented for future estimation. Generally, autoencoders have the ability to compress and reconstruct features. Among them, the conditional variational autoencoder (CVAE) \cite{cvae, vae} can generate corresponding feature vectors or matrices based on different conditions. \textbf{Therefore, our first motivation lies in using CVAE to generate transfer vectors that have both randomness and directionality, as a supplement to the original state transition data.} We construct a state transition autoencoder (STA) based on CVAE for inference.

STA solves the reasoning problem, but we need conditions to perform reasoning. Here, the conditions should be actions that have not occurred, also known as counterfactual actions. And, the counterfactual action will produce a counterfactual result, this process is called counterfactual (or virtual) state transition. Generally speaking, actions that have already occurred are known, but there may be a very large number of actions that have not occurred and are difficult to traverse. \textbf{In this case, our second motivation lies in achieving efficient and resource-saving counterfactual action sampling.} Simply put, we construct a Gaussian kernel density estimation function in the action space and determine representative counterfactual actions through the method of maximum entropy sampling.

Traditionally, building a model to fit the environment is a good way to predict future states and reward values, as implementation by Janner et al. \cite{MBPO}, but we found that in many environments, the prediction error of reward values is very large. \textbf{Thus, our third motivation is to reduce this error, by matching the real reward values to counterfactual state transitions.} After that, we can construct complete counterfactual experiences. Additionally, this matching method is based on the assumption of bisimulation, which will be introduced in Section \ref{Background}.


In short, by integrating the methods proposed previously, we use the counterfactual actions obtained from sampling as conditions, enabling the STA inference to obtain possible counterfactual results. We then compare the counterfactual result with real data, pairing the closest data through the minimum Euclidean distance and assigning the true reward value to the counterfactual result. Our concept of counterfactual is consistent with Pearl's work \cite{CausalInference}. We present an explanatory diagram of counterfactual reasoning learning in Fig. \ref{CEA_figure}.

In summary, this paper introduces the STA, and counterfactual experience augmentation (CEA) algorithm. The code is available in \url{https://github.com/Aegis1863/CEA}.

Our main contributions are as follows:


\begin{itemize}
	\item[1.] We introduce the STA, a model that learns state transition dynamics with inherent randomness, enabling counterfactual reasoning capabilities.
	\item[2.] We design a maximum entropy sampling optimization method based on Gaussian kernel density estimation for counterfactual action sampling.
	\item[3.] We propose a comprehensive CEA algorithm, which uses counterfactual action sampling and STA model for counterfactual reasoning to alleviate the OOD problem of learning data and can effectively improve the performance of the backbone model.
\end{itemize}

In the following text, Section \ref{relatedwork} introduces related work, while Section \ref{Background} introduces basic methods. Our implementation is in Section \ref{Methodology}, which also includes some experimental demonstrations. Section \ref{Metrics} introduces the testing environment and comparison algorithms. We present the experimental results and analysis in Section \ref{Experiment}, and provide the details of the innovation points in Section \ref{Innovation}. Finally, we give our discussion and conclusion in Section \ref{discussion} and \ref{conclusion}.

\section{Related works} \label{relatedwork}

Our work belongs to model-based RL. We also utilize the theory of causal inference and bisimulation as the basis for data augmentation. Therefore, this section presents this existing research.

\subsection{Deep RL}

Q-learning \cite{Watkins1992}, a fundamental off-policy technique, estimates cumulative rewards for state-action pairs without requiring an environment model. The introduction of Deep Q Networks (DQN) \cite{Mnih2015} enhanced this approach by using convolutional neural networks to approximate Q-values, incorporating experience replay and target networks for improved stability. Deep Deterministic Policy Gradient (DDPG) \cite{lillicrap2019continuous} extends these principles to continuous action spaces, using an actor-critic architecture with off-policy learning. Soft Actor-Critic (SAC) \cite{haarnoja2019soft, christodoulou2019soft} further improves this framework by adding entropy regularization to balance exploration and exploitation. Experience management, through techniques like prioritized experience replay (PER) \cite{schaul2016prioritized}, improves data efficiency by prioritizing valuable experiences. At the same time, some multi-agent solutions based on Deep RL have gradually been proposed \cite{10311540, 10534267, MAPPO}.

\subsection{Model-based and causal RL}

Model-based RL leverages environment models to enhance decision-making. The probabilistic ensemble trajectory sampling algorithm \cite{NEURIPS2018PETS} uses a probabilistic model to handle uncertainty, while other similar methods \cite{hafner2020dream, MBPO, hafner2019learning} also show strong performance.

Causal RL, which integrates causal inference into learning, is an emerging field. Yang et al. \cite{YANG2023243} introduced a causal factor for agents in multi-agent tasks, while Sun et al. \cite{Sun2024} used adversarial models to augment experiences, though both lacked clear solutions for reward signals in counterfactual state transitions. Chao et al. \cite{Chao2021} proposed using structural causal models for state dynamics, providing a more detailed theoretical approach. We noticed that Pitis et al. \cite{mocoda} proposed a model-based causal RL method, MoCoDa, which is combined with data augmentation.

\subsection{Data augmentation RL}

As mentioned in the introduction, many data augmentation RL serves tasks with observation data as images, because it is easy to draw on the experience of computer vision. They have some methods to supplement data using cropping, rotation, and noise, such as RAD \cite{RAD},  DrQ \cite{drq}. While CURL \cite{curl}, DBC \cite{dbc} and SODA \cite{SODA} improve the model recognition ability through contrastive learning. We discovered a new data augmentation method based on vector observations, namely Acamda \cite{Sun2024} proposed by Sun et al. Acamda proposes to use bidirectional conditional causal generative adversarial network for reasoning, which is similar to ours. Unfortunately, this method can hardly be reproduced. We provide a horizontal comparison of these peer works in Table \ref{table:compare}.

\begin{table}[t]
    \centering
	\caption{Comparison of data augmentation RL. In the table, Causal indicates whether the method involves a causal framework, Domain represents the observed form to be addressed. Model denotes whether the method is model-based. Year corresponds to their proposed time. Code indicates whether the method has open-source code.}\label{table:compare}
    \begin{tabular}{lllllll}
    \toprule
        Algorithm             & Causal & Domain  &  Model  & Year      & Code \\
		\midrule
		MBPO \cite{MBPO}	  & no     & any     &   yes     & 2019   & yes  \\
        RAD \cite{RAD}        & no     & pixel   &   no      & 2020   & yes  \\
        CURL \cite{curl}      & no     & pixel   &   no      & 2020   & yes  \\
		DrAC \cite{DrAC}	  & no     & pixel   &   no      & 2020   & yes  \\
        DrQ \cite{drq}        & no     & pixel   &   no      & 2021   & yes  \\
        SODA \cite{SODA}      & no     & pixel   &   no      & 2021   & yes  \\
		MoCoDa \cite{mocoda}  & yes    & pixel   &   yes     & 2022   & yes  \\
        Acamda \cite{Sun2024} & yes    & vector  &   yes     & 2024   & no   \\
		\midrule
        CEA (Ours)            & yes    & vector  &   yes      & 2024   & yes  \\
	\bottomrule
    \end{tabular}
\end{table}

Next, we will introduce the preliminary theory and method of our work.

\section{Preliminary}\label{Background}

This section introduces preliminary methods, which are the foundation of our model.

\subsection{Reinforcement learning} \label{rlconcept}

RL are usually defined in Markov decision processes (MDP), as detailed by Sutton et al. \cite{Sutton1988,sutton1999,sutton2018reinforcement}. We assume that the underlying environment is a MDP, characterized by the tuple $\mathcal{M} = (\mathcal{S}, \mathcal{A}, \mathcal{P}, \mathcal{R}, \gamma)$. In this framework, $\mathcal{S}$ denotes the set of states, $\mathcal{A}$ represents the set of actions, $\mathcal{P}(s'|s, a)$ specifies the probability of transitioning from state $s \in \mathcal{S}$ to state $s' \in \mathcal{S}$, and $\gamma \in [0, 1)$ is a discount factor that determines the importance of future rewards. An agent chooses actions $a \in \mathcal{A}$ according to a policy $a \sim \pi(s)$, which then influences the next state $s' \sim \mathcal{P}(s, a)$ and generates a reward $r = \mathcal{R}(s) \in \mathbb{R}$. The agent's goal is to find a policy that maximizes the expected sum of discounted rewards over time: \smash{$\max_{\pi}\mathbb{E}_\mathcal{P}[\sum_{t=0}^{\infty}\gamma^t\mathcal{R}(s_t)]$}. The current general deep RL SOTA algorithms all serve the goal of maximizing the final reward, and the main method includes temporal difference and policy gradient.

CEA are built on DQN and DDPG, which have become important application methods of temporal difference. We will give a brief introduction below.

Considering that the timing difference error is usually expressed as:
\begin{equation}\label{tderror}
	\delta_t=r_{t}+\gamma\cdot\max_{a'}Q(s_{t+1},a')-Q(s_t,a_t).
\end{equation}

The parameters of the DQN are updated as
\begin{equation}\label{eq:ddqn}
	\begin{aligned}
		\omega' = \mathop{\operatorname{argmin}} \limits_{\omega} \frac{1}{2N} \sum^N_{i=1}\Big[Q_{\omega}(s^i_{t},a^i_{t}) - \big( r_t^i + \gamma Q_{\omega}(s^i_{t+1}, \mathop{\operatorname{argmax}} \limits_{a} Q_{\omega}(s^i_{t+1},a))\big) \Big]^2,
	\end{aligned}
\end{equation}
where $\omega$ is network parameters; $Q_{\omega}$ is the Q function; $s^i_t$, $a^i_t$, $r^i_t$ and $s^i_{t+1}$ respectively stand for the state, action, reward and next state in time step $t$ or $t+1$; $N$ is the number of samples. While $\gamma$ is the discount factor.

In the neural network model, we modify the structure to have a dueling form. The optimal target is defined as:
\begin{equation}
	Q_{\lambda,\alpha,\beta}(s,a)=V_{\lambda,\alpha}(s)+A_{\lambda,\beta}(s,a)-\frac{1}{|A|}\sum_{a'}A_{\lambda,\beta}\left(s,a'\right),
\end{equation}
where $V_{\lambda,\alpha}(s)$ is the state value function, $A_{\lambda,\beta}(s,a)$ is the advantage function. $\lambda$ represents the sharing parameters of $V$ and $A$ function. $\alpha$ and $\beta$ are the private parameters of $V$ and $A$ function respectively. The purpose of subtracting the average value from the advantage function is to guide the network to update the value function $V$. Otherwise, the network may be inclined to make $V=0$ and $A$ degenerate to $Q$. On the other hand, calculating the difference between $A$ and its average value is also to allow the neural network to learn the value difference between various actions.

DQN can usually only cope with discrete action decision tasks. In order to extend the value learning method to the continuous action space, DDPG constructs two networks, one serving as the actor and the other as the critic. The actor network is based on a policy function and directly outputs a deterministic action for a given state, which differentiates it from traditional value-based methods that output a probability distribution over actions. The critic network, on the other hand, is based on the $Q$-value function and evaluates the value of a given state-action pair. In DDPG, the actor network relies on the critic network to optimize its policy. Specifically, the critic network learns the $Q$-values by minimizing the temporal difference error, while the actor network updates its policy by maximizing the $Q$-values provided by the critic network. This combination of policy gradient methods and value-based methods allows DDPG to efficiently learn policies in continuous action spaces.

In simple terms, the update formula of Q function can be expressed as:
\begin{equation}
	Q(s_t, a_t) \leftarrow r_t+ \gamma\max\limits_{a^*} Q(s_{t+1}, a^*).
\end{equation}

In off-policy, a batch of experiences is usually extracted from the experience pool as a small batch. Each experience consists of a tuple $\{s, a, s', r\}$, and the Q value is estimated by the mean calculation method:
\begin{equation}
	Q(s, a) \approx \frac{1}{N}\sum_{i=1}^Nr_i+ \gamma\max\limits_{a^*} Q(s'_i, a^*),
\end{equation}
where $s'$ is the random variable form of $s_{t+1}$. Therefore, increasing $N$ is conducive to accurately estimating the $Q$ function and reducing the estimation variance:
\begin{equation}
	\mathrm{Var}(\hat{Q})\propto\frac{\sigma^2}N,
\end{equation}
and this is equivalent to alleviating the OOD of state distribution.

Based on traditional off-policy algorithm, PER takes into account the importance of experience.

\subsection{Prioritized experience replay}

PER assigns an importance value to each experience, which is directly proportional to the temporal difference error as computed by Equation \ref{tderror}. The higher the error, the greater the assigned importance.  The importance weight update process is governed by:
\begin{equation}\label{eq:per}
	\begin{aligned}
		P_i & =\frac{\omega_i^\delta}{\sum_k\omega_k^\delta},   \\
		w_i & =\left(\frac{1}{N}\cdot\frac{1}{P_i}\right)^\eta,
	\end{aligned}
\end{equation}
where $\delta$ determines the priority level, and it degrades to uniform sampling when $\eta$ is 0. The size of experience pool is denoted by $N$. $\eta$ stands for a parameter controlling the weight magnitude, typically increases gradually during the training process. We consider PER to manage numerous experiences and ensure that low-value samples are underestimated. However, PER may not always help, which will be used as an ablation condition in the experiments.

We next introduce the concept of counterfactual inference.

\begin{wrapfigure}[15cm]{r}{5cm}
	\centering
	\includegraphics[width=0.7\linewidth]{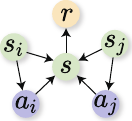}
	\caption{Causal directed acyclic graph of state transition process. In this case, different state-action pairs will lead to the same result, satisfying the bisimulation property.}\label{DAG}
\end{wrapfigure}

\subsection{Counterfactual inference}

Giving Fig. \ref{DAG} as directed acyclic causal graph, $s_i$ and $s_j$ are two different states, $a_i$ and $a_j$ are two corresponding actions. Different state-action pairs may lead to the same or similar state transition result $s$ and the same reward $r$. This is consistent with the bisimulation property mentioned in the previous part.

In the figure, $s_i$ and $a_i$ affect the intermediate state $s$, and $s$ further affects the reward $r$. $s$ can be regarded as the mediating variable of $s_i$ and $a_i$ on $r$. This constitutes a case of the front door effect. The same is true for $s_j$ and $a_j$. Therefore, $s$ can be used to analyze the causal effect of different action states on the final reward $r$. For example, we consider randomly sampling some $s$ dividing into two groups, $s_i$ and $s_j$, and find the $a_i$ corresponding to $s_i$. Then sampling some not occurred $\hat{a_j}$ by maximum entropy method, and using the autoencoder to generate $s$. $\hat{a_j}$ represents a counterfactual action that the agent has not yet practiced.

Next, we present the bisimulation property and why we emphasize this point.

\subsection{Bisimulation}

We use bisimulation as a theoretical basis because obviously no model can predict the future accurately, but in some cases, it can be approximated, which depends on the bisimulation property. At the same time, we also need to point out that not all tasks meet this condition. We will define bisimulation based on the definition of MDP in Subsection \ref{rlconcept}. Bisimulation is a form of state abstraction that groups states $s_i$ and $s_j$ that are "behavioral equivalent" \cite{li2006towards}.

For a standard definition \cite{givan2003equivalence}: an equivalence relation $B$ between states is a bisimulation relation if, for all states $\mathbf{s}_i, \mathbf{s}_j \in \mathcal{S}$, that are equivalent under $B$ (denoted $\mathbf{s}_i \equiv_B \mathbf{s}_j$) the following conditions hold:
\begin{equation}
	\begin{aligned}
		\mathcal{R}(\mathbf{s}_{i},\mathbf{a}) & = \mathcal{R}(\mathbf{s}_{j},\mathbf{a})\quad\forall \mathbf{a}\in\mathcal{A}, \\\mathcal{P}(G\mid\mathbf{s}_{i},\mathbf{a})&= \mathcal{P}(G\mid\mathbf{s}_{j},\mathbf{a})\quad \forall \mathbf{a}\in\mathcal{A},~\forall G\in \mathcal{S}_{B},
	\end{aligned}
\end{equation}
where $\mathcal{S}_B$ is the partition of $\mathcal{S}$ under the relation $B$ (the set of all groups $G$ of equivalent states), and $\mathcal{P}(G|\mathbf{s},\mathbf{a})=\sum_{s'\in G}\mathcal{P}(\mathbf{s}'|\mathbf{s},\mathbf{a})$.

\textbf{In short, for two different states, if the actions they take lead to the same state transition results, they can be regarded as satisfying the bisimulation property.} However, in the continuous state space, it is almost impossible for two states to be exactly the same, so we regard the case of similar state distribution as approximately satisfying the bisimulation property.

In our work, if the result of state transfer in virtual experience is the same as the result of a real state transfer, then we think their state transfer reward is the same.

The maximum entropy sampling, STA and other important implementations will be introduced in the next section.

\section{Methodology} \label{Methodology}

This section introduces all implementations of our work.

\subsection{State transition autoencoder}

RL based on generative models has made great progress, such as diffusion RL and generative adversarial RL \cite{Ho2016GenerativeAI,ICML19atorabi,wang2023diffusioninrl}. This paper presents a simple and practical method based on a CVAE.

In order to make the model focus on more significant characteristics, STA learns the difference between state transitions and represented by $d$. Let $d$ as an input and $d'$ is a generated output. $d$ is calculated as:
\begin{equation}
	d_t = s_{t+1} - s_{t},
\end{equation}
where $s_{t+1}$ and $s_{t}$ are next and current state, respectively. The way to get the agent to learn state transitions is a classic application in \cite{bansal2017} and \cite{MBPO}, which emphasizes understanding the underlying transition process, making it more amenable to learning.

To incorporate the counterfactual action $\hat{a}$ as conditions in the latent variable, the process differs depending on the type of action space. In discrete action space, the actions are countable, all actions other than those that have been performed are counterfactuals. For continuous action spaces, sampling is required, and the number of samples for $\hat{a}$ should be manually adjusted. Counterfactual action sampling in continuous action space will be introduced in the next subsection

Consequently, STA follows the structure of CVAE, focusing on the dynamics of state transitions. By introducing a latent variable $z$, we can compute the probability density function of $d$ conditioned on $a$ through the construction of marginal likelihood:
\begin{equation}\label{eq:likelihood}
	\log p(d|a)=\log\int p(d|z,a) p(z|a)dz.
\end{equation}

We introduce the variational distribution $q(z|d,a)$ as an approximate representation, aiming to capture the underlying structure of the latent variables $z$ given the observed data $d$ and condition $a$. It should be noted that we hope that $p(z|a)$ conforms to the standard Gaussian distribution, which is convenient for calculation and sampling.

According to Jensen's inequality, we can derive evidence lower bound:
\begin{equation}
	\begin{aligned}
	\log p(d|a) &\geq \mathbb{E}_{q(z|d,a)}\left[\log\frac{p(d|z,a)p(z|a)}{q(z|d,a)}\right]\\
		        & =\mathbb{E}_{q(z|d,a)}\left[\log p(d|z,a)\right]-\\
				& \quad \mathbb{D}_{KL}\left[q(z|d,a)||p(z|a)\right],
	\end{aligned}
\end{equation}
where the objective involves a reconstruction loss term $\mathbb{E}_{q(z|d,a)}[\log p(d|z,a)]$, which measures the discrepancy between the generated and real data. While $\mathbb{D}_{KL}$ is the Kullback-Leibler divergence, which quantifies the difference between the approximate distribution $q(z|d,a)$ and the prior distribution $p(z|a)$.

We mentioned that $p(z|a)$ is a standard Gaussian distribution, and $q(z|d,a)$ also follows a Gaussian distribution, but its mean $\mu$ and variance $\sigma^2$ are determined by an encoder model (a neural network). At this time, $\mathbb{D}_{KL}\left[q(z|d,a)||p(z|a)\right]$ is calculable, that is
\begin{equation}
	\mathcal{L}_{kl}=\frac{1}{2} \sum_{i=1}^n\Big[\mu_{i}^2 + \sigma_{i}^2 - \log\sigma_{i}^2 - 1\Big],
\end{equation}
where $i$ refers to the $i$th sample. We refer to the proof of Su \cite{kexuefm5253} for details provided in Appendix \ref{ap:proof1}.

The reconstruction error $\mathbb{E}_{q(z|d,a)}\left[\log p(d|z,a)\right]$, computed as the mean squared error, is the difference between the reconstructed data $d'_i$ and the original data $d_i$, and is given as:
\begin{equation}
	\mathcal{L}_{r}=\frac{1}{2n} \sum_{i=1}^n\Big(d'_i - d_i\Big)^2.
\end{equation}

Therefore, the optimization goal of STA is to minimize the total loss:
\begin{equation}
	\mathcal{L}=\mathcal{L}_{r} + \mathcal{L}_{kl}.
\end{equation}

By minimizing the loss, generated $d'$ is optimized to closely approximate $d$. With the use of STA, we proceed by sampling a vector $z'$ from a Gaussian distribution, concatenating it with counterfactual actions and forwarding it to the decoder. This process generates a transition vector $d'$. Subsequently, we create a counterfactual next state by adding $d'$ to the current state $s$. The next challenge lies in providing an appropriate virtual reward signal, which we will address in the subsequent subsection.

\begin{wrapfigure}[38cm]{r}{0.5\linewidth}
	\centering
	\subfigure[Sampling in 1-dimensional action space. Three new samples that can be optimized are set for one initial sample.]{
		\includegraphics[width=\linewidth]{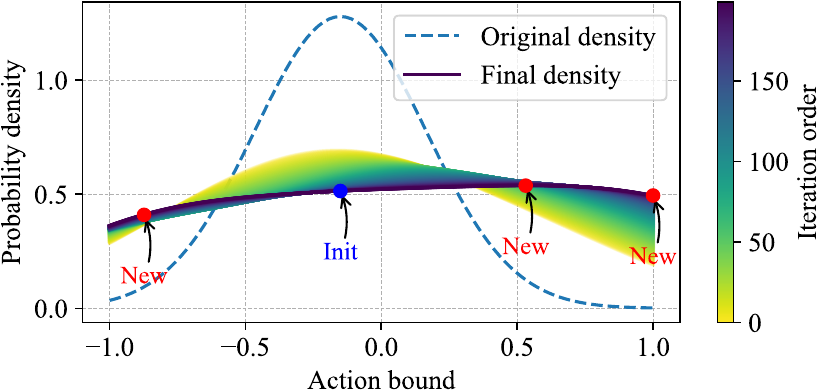}
		\label{kde1d}
	}
	\subfigure[Sampling in 2-dimensional action space. Nine new samples that can be optimized are set for one initial sample.]{
		\includegraphics[width=\linewidth]{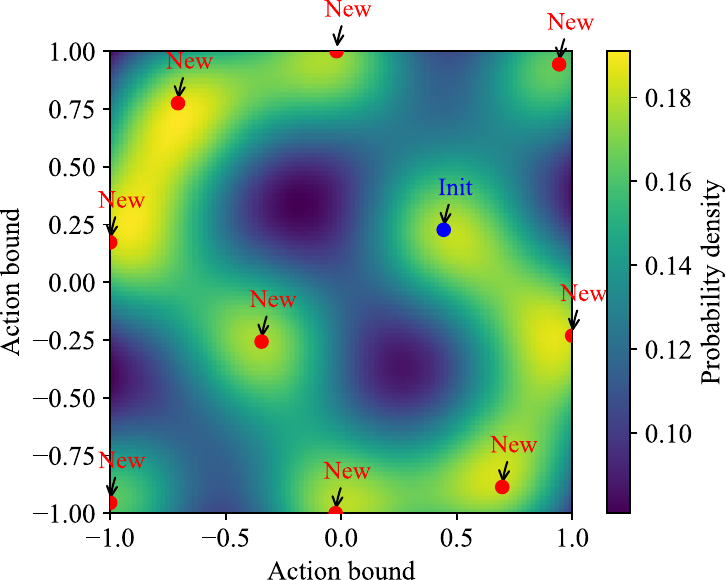}
		\label{kde2d}
	}
	\caption{Sampling based on kernel density estimation} \label{kde}
\end{wrapfigure}

\subsection{Counterfactual action sampling}

Action sampling actions from discrete action spaces are easy, the method used in this paper is to reason about all unsampled action choices as counterfactual actions. Therefore, this section introduces how to extend CEA to the counterfactual action sampling method in continuous action spaces. This approach can be applied to discrete action spaces as well.

In continuous action spaces, the sampling is troublesome. This paper proposes to construct a probability density based on existing data through Gaussian kernel density estimation, then randomly initialize some samples. These randomly initialized samples are the counterfactual action samples to be optimized.

The reason for choosing the Gaussian kernel function is that it has no parameter and has an explicit expression. At this point, we can focus only on optimizing the input variables without having to tune the model. While the explicit expression makes the optimization target formula differentiable, so that the approximate solution can be obtained by the gradient descent. Another benefit is that the Gaussian kernel gives smoother estimates and can more easily handle data with long-tailed distributions, which aligns our data characteristics.

The optimization of these counterfactual samples should maximize the entropy of the new kernel density. \textbf{In other words, we hope that counterfactual samples can be fully distributed in those desolate spaces.} The following is the specific sampling method.

We initialize some counterfactual actions $\hat{a}$ and merge it with the known samples $a$ to get a new action set $A$. While it is theoretically possible to have multiple $a$, in practice there is only one, because we will deal with each experience sample separately, with only one $a$ in each experience.

Our goal is
\begin{equation}\label{eq:optimize_a}
	\max_{\{\hat{a}_1,\hat{a}_2,\ldots,\hat{a}_n\}}-\sum_{x\in A}p(x)\log p(x).
\end{equation}

We implement gradient descent method for numerical optimization. Let $y(x) = p(x)\log p(x)$ and be shortened to $y$. Split the domain into $m$ parts and use trapezoidal integration to approximate the entropy:
\begin{equation}
	\begin{aligned}
		H(x) & = -\int p(x)\log p(x) \,\mathrm{d}x                                      \\
		     & \approx -\sum_{j=1}^{m-1}\left(\frac{x_j-x_{j-1}}{2}\right)(y_j+y_{j-1}).
	\end{aligned}
\end{equation}

According to the Gaussian kernel density estimation, for $p$, we have
\begin{equation}
	p(x) = \frac{1}{|A|h\sqrt{2\pi}}\sum^{|A|}_{i=1}\exp\left\{-\frac{(x-a_i)^2}{2h^2}\right\},
\end{equation}
where $|A|$ represents the length of the set $\hat{A}$. The gradient of $H$ w.r.t $\hat{a}$ is
\begin{equation}\label{opttarget}
	\frac{\partial  H}{\partial \hat{a}} = -\sum_{j=1}^{m-1}\left(\frac{x_j-x_{j-1}}{2}\right)\left(\frac{\partial y_j}{\partial \hat{a}}+\frac{\partial y_{j-1}}{\partial \hat{a}}\right).
\end{equation}

Since the value of $j$ depends on $m$, we remove the subscript and consider the expression of $y$ alone:
\begin{equation}
	\begin{aligned}
		 & y(x) = p(x)\log p(x),   \\
		 & \log p(x) = -\log(|A|h\sqrt{2\pi}) + \sum_{i=1}^{|A|}\left[-\frac{(x-a_i)^2}{2h^2}\right],  \\
		 & \frac{\partial y}{\partial \hat{a}} = \frac{\partial p(x)}{\partial \hat{a}}\big[\log p(x)+1\big],
	\end{aligned}
\end{equation}
where
\begin{equation}
	\begin{aligned}
		\frac{\partial p(x)}{\partial \hat{a}}
		 & = -\frac{1}{|A|h\sqrt{2\pi}}\exp\left\{-\frac{(x-\hat{a})^2}{2h^2}\right\} \cdot \frac{x-\hat{a}}{h^2} \\
		 & = -\frac{x-\hat{a}}{|A|h^3\sqrt{2\pi}}\exp\left\{-\frac{(x-\hat{a})^2}{2h^2}\right\}.
	\end{aligned}
\end{equation}

Therefore, the differential of $y$ w.r.t $\hat{a}$ can be found as
\begin{equation}\label{yadifferential}
	\begin{aligned}
		\frac{\partial y}{\partial \hat{a}} & = \frac{\partial p(x)}{\partial \hat{a}}\big[\log p(x)+1\big]                              \\
		                                    & = -\frac{x-\hat{a}}{|A|h^3\sqrt{2\pi}}\exp\left\{-\frac{(x-\hat{a})^2}{2h^2}\right\} \cdot \\ &\left\{-\log(|A|h\sqrt{2\pi}) + \sum_{i=1}^{|A|}\left[-\frac{(x-a_i)^2}{2h^2}\right] + 1\right\}.
	\end{aligned}
\end{equation}

The calculation of $\smash{\frac{\partial y_{j}}{\partial \hat{a}}}$ and $\smash{\frac{\partial y_{j-1}}{\partial \hat{a}}}$ is similar. We can substitute them into Equation \ref{opttarget} to calculate the specific gradient. In summary, the gradient of the initial sampling action based on maximizing information entropy is calculable. Once the gradient value is calculated, it can be multiplied by a learning rate and added to the original values of $\hat{a}$ for multiple iterations, as shown in Equation \ref{opteq}. Therefore, those samples can be optimized.
\begin{equation}\label{opteq}
	\begin{aligned}
		\hat{a} \leftarrow \hat{a} + lr \cdot \frac{\partial  H}{\partial \hat{a}}.
	\end{aligned}
\end{equation}

In this case, Equation. \ref{eq:optimize_a} is solvable, therefore, here we provide the expression for obtaining the counterfactual action $\hat{a_i}$:
\begin{equation}\label{eq:getcunterfacter_a}
	\{\hat{a}_1,\hat{a}_2,\ldots,\hat{a}_n\} = \mathop{\mathrm{argmax}}\limits_{\hat{a}_1,\hat{a}_2,\ldots,\hat{a}_n}-\sum_{x\in A}p(x)\log p(x),
\end{equation}
where the set $A$ include both a real action $a$ and several counterfactual actions $\hat{a}_1,\hat{a}_2,\ldots,\hat{a}_n$. We only optimize those counterfactual actions.

Fig. \ref{kde1d} and Fig. \ref{kde2d} are the sampling experimental results of this method in one-dimensional and two-dimensional data. A known sample is marked with \textit{Init}, and newly added points are marked with \textit{New}. The initial positions of \textit{New} are randomly, and its final position is the result of iterative optimization. It can be observed that the final positions of \textit{New} are evenly distributed in areas where there were originally no samples.

Next, we introduce how to construct a complete counterfactual experience.

\subsection{Closest transition pair} \label{CTP}

Here, we will present the framework of CEA. We commence by constructing an MDP to elucidate
the concept of counterfactual actions, followed by detailing the algorithmic procedure of CEA.

\begin{figure}[htbp]
	\centering
	\includegraphics[width=0.6\linewidth]{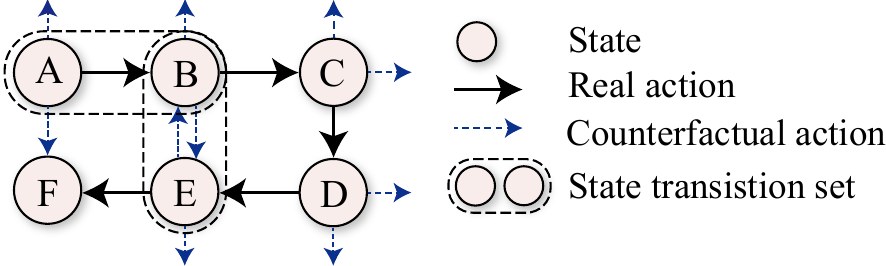}
	\caption{MDP with counterfactual actions}\label{counterfactualMDP}
\end{figure}

Fig. \ref{counterfactualMDP} illustrates a MDP example with both a real track and counterfactual actions, and fully satisfies the bisimulation property. Assuming action space is discrete with 3 action options, as shown by the arrows in Fig. \ref{counterfactualMDP}. This track consists of 7 stages from A to F, with 3 discrete action options per stage. The solid arrows represent real actions, leading to the next state, while dashed arrows symbolize counterfactual actions. At state E, the agent chooses to transfer to F. Using STA, we can infer transition vectors conditioned on hypothetical actions, allowing us to compute an estimated next state.  If, for instance, E could hypothetically reach B through a counterfactual action, we term \{A, B\} and \{E, B\} as closest transition pair (CTP). Therefore, as long as the next state transferred to is consistent or similar, a CTP can be formed. Since \{A, B\} is a factual history with complete information, we assign the reward of \{A, B\} to \{E, B\}. In simple terms, the implementation of CTP can be expressed as finding index $i^*$ of samples:
\begin{equation}\label{eq:ctp}
	\begin{aligned}
		 & r^*=r_{i^*}, \\&i^*=\arg\min_id(\hat{s}_i,s),
	\end{aligned}
\end{equation}
where $\hat{s}$ is the counterfactual next state calculated by STA, and $s$ is the real state that already exists in the experience pool. The distance function $d$ can use Euclidean distance or Manhattan distance.

This shows the result orientation of bisimulation. We give a direct example, with the abuse of symbols, which are only used here. Assume that we known state transition pairs such as $\{s^r_t, a^r_t, s^r_{t+1}, r\}$, which represent the state and action in time step $t$, and the next state and reward of this state transition. Meanwhile, we give a counterfactual state transition pair $\{s^c_t, a^c_t, s^c_{t+1}\}$ (It currently lacks $r$). If the transition results $s_{t+1}$ of the two are very close, that is, $s^r_{t+1} \approx s^c_{t+1}$, then we believe that the reward values of the two state transition pairs are consistent. At this time, we copy $r$ in $\{s^r_t, a^r_t, s^r_{t+1}, r\}$ to $\{s^c_t, a^c_t, s^c_{t+1}\}$. Then, we construct a complete counterfactual experience $\{s^c_t, a^c_t, s^c_{t+1}, r\}$.

\subsection{Counterfactual experience augmentation} \label{Counterfactual-experience}

\begin{wrapfigure}[19cm]{r}{8cm}
	\centering
	\includegraphics[width=\linewidth]{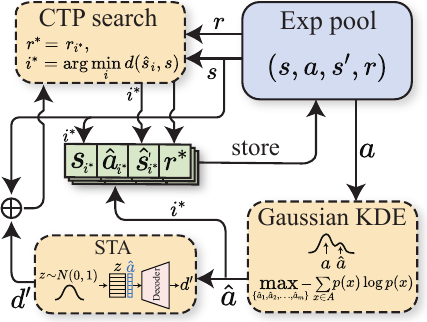}
	\caption{CEA framework}\label{CEA_framework}
\end{wrapfigure}

Since the counterfactual next state cannot be completely consistent with the real state when the state is in continuous spaces, we measure the distance and only consider those state groups with close distance as a CTP. At the same time, the randomness of the autoencoder also helps to simulate the possible non-stationarity of the environment.

As outlined in Algorithm \ref{alg:CEA}, it should be noted that STA needs to be trained in advance, and its training data only needs to be randomly collected from an environment. We compute and sort the difference between the counterfactual next state and the real next state. A threshold ratio can be given as a hyperparameter to filter out those most similar pairs and share the real reward with the counterfactual one. The ratio can also be changed to a specific distance, but it increases the difficulty of fine-tuning. The algorithm framework diagram is shown in Fig. \ref{CEA_framework}.

By incorporating rewards into counterfactual experiences, these expanded experiences are added to the experience pool and sampled alongside real ones. The significance of PER lies in managing the large number of potential counterfactual experiences, as with a discrete action space of $m$ and $n$ real experience samples, there can be $n(m-1)$ counterfactual transitions, leading to rapid expansion of experience. While the similar situation occurs in the continuous action space.

CEA is typically employed in off-policy learning, which is performed before the agent samples the experience. CEA is not supposed to execute frequently. Otherwise, there will be a lot of redundant and similar counterfactual experiences.

It should be noted that the CEA method is built on the basis of Rainbow DQN (RDQN) and DDPG, which are the backbones corresponding to discrete action space tasks and continuous action space tasks respectively. After introducing our methods, the next section gives experimental indicators, details and scenarios.
\begin{algorithm}[htbp]
	\caption{Counterfactual experience augmentation}\label{alg:CEA}
	\begin{algorithmic}[1]
		\State \textbf{Input:} replay buffer, STA
		\State \textbf{Output:} Augmented replay buffer
		\State Randomly sampling $\mathbf{unsampled}$ real experience $(s, a, r, s')$ from replay buffer
		\State Sampling counterfactual actions $\hat{a}$ from Equation. \ref{eq:getcunterfacter_a}
		\State Calculate counterfactual state transition vectors $d$ by giving $\hat{a}$ to STA from $p(d|z,a)$ mentioned in Equation. \ref{eq:likelihood}
		\State Repeat each state times to create $expand\_s$ to align the number of counterfactual data
		\State Compute estimated next states: $\hat{s'} = expand\_s + d$
		\State Retrieve $\mathbf{real}$ experiences from replay buffer as $(all\_s, all\_a, all\_r, all\_ns)$
		\State Compute distances between $all\_ns$ and $\hat{s'}$ from Equation. \ref{eq:ctp} and pair the closest top parts and share the real reward $r^*$ to the virtual one
		\State Combine $(expand\_s, \hat{a}, \hat{s'}, r^*)$ as counterfactual experiences
		\State Store counterfactual experiences in replay buffer\\
		\Return replay buffer
	\end{algorithmic}
\end{algorithm}

\section{Experiment settings} \label{Metrics}
We introduce both discrete and continuous action space experimental environments. They are all based on the standard environment wrapped by gymnasium \cite{towers2024gymnasium}.

For discrete action space environments, we use SUMO and Highway, both of which are challenging traffic simulation environments. And they have relatively suitable bisimulation properties. The former is a professional open-source traffic simulation system, and the latter is a simplified driving decision environment. Due to the rapid changes in traffic environments, they are inherently non-stationary \cite{Alegre2021,Padakandla2020}.

For the continuous action space environments, we use the standard benchmarks Pendulum and Lunar lander (continuous control). They participated in almost all RL algorithm benchmarks. Specific task content and environmental parameters refer to their official documentation \cite{towers2024gymnasium}.

For SUMO and Highway, since they are not widely used simulation benchmarks in the field of RL algorithms, we will give a brief introduction.

\begin{figure}[htbp]
	\centering
	\subfigure[SUMO traffic light control. An Agent will make decisions for a traffic light at an intersection, and the higher the number of waiting vehicles at the intersection, the higher the reward. Its actions are a finite combination of traffic light configurations.]{
		\includegraphics[width=0.48\textwidth]{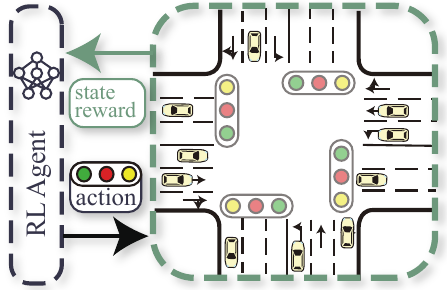}\label{sumo-demo}
	}
    \hspace{0.5cm}
	\subfigure[Highway driving decision. Cars are driving on the expressway. The green car, as an agent, needs to make lane-changing decisions. The higher the reward will be when collisions are minimized and the speed is as fast as possible. Its action space size is 4, which are acceleration, deceleration, left turn, or right turn.]{
		\includegraphics[width=0.44\textwidth]{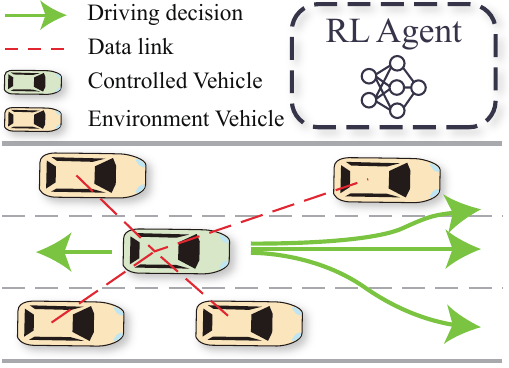}\label{highway-demo}
	}
	\caption{Discrete action space benchmark, SUMO and Highway} \label{sumo-highway}
\end{figure}

\subsection{SUMO}

SUMO, an established professional traffic simulation system \cite{SUMO2018}, is a popular choice in the traffic simulation research community \cite{sumorl,ault2021reinforcement,SUMOearly}. We believe that improvements here will be of significant help and inspiration to the field. We employ SUMO as one of our benchmark environment for traffic signal control, as depicted in Fig. \ref{sumo-demo}. The goal of the algorithm is to control traffic signals to avoid congestion at the intersection. Given the discrete nature of traffic signal phases, the action space is finite. We analyze the training and control performance of various algorithms using the total reward to evaluate the effectiveness of their training. Total reward is called return as well. Reward in each step is defined as:
\begin{equation}\label{td-reward}
	r_t = w_{t+1} - w_{t},
\end{equation}
where $w_{t+1}$ is sum of waiting times of all vehicles at intersection at time $t+1$, and the same is true for $w_t$. Therefore, this reward encourages a smaller difference in the total waiting time at current and previous time points. In fact, other reward signals such as the number of vehicles in queue can also be used, because the fewer vehicles in the queue, the stronger the algorithm's ability to optimize traffic, but experiments show that the reward signal defined in Equation \ref{td-reward} performs best. The state design follows the design of Alegre et al. \cite{Alegre2021}, which mainly includes lane density, queue vehicle density, etc.

\subsection{Highway}

Highway \cite{highwayenv} is a simplified road environment, including various scenarios like highways, roundabouts, ramps, etc. It has many lane-changing studies \cite{yanjun, tian2024enhancingautonomousvehicletraining} and has become a new test benchmark in the fields of general and transportation. The difference from SUMO is that the control subject is the vehicle instead of the traffic light. Schematic diagram of the two tasks refer to Fig. \ref{highway-demo}. It simulates a small car on a multi-lane highway. Other vehicles on the road are randomly generated, so the control goal of the algorithm is driving decisions, such as which direction to avoid, overtake or slow down. The action space is discrete and simplified to four options: accelerate, slow down, left turn, and right turn.

A controlled vehicle can observe its own and four near vehicles. The observation vector of each vehicle, includes the position, speed and other information in the environment. The reward function is defined as:
\begin{equation}\label{highway-reward}
	r_t=\phi \frac{v_t-v_{\min}}{v_{\max}-v_{\min}}-\kappa \text{ collision}_t,
\end{equation}
where $v_t$, $v_{\text{min}}$, and $v_{\text{max}}$ are the current speed, minimum speed, and maximum speed, respectively. For the first term, it encourages the vehicle to keep high speed to learn to avoid the car in front and gives a reward weight $\phi$. Collision is a binary variable that indicates whether a collision occurs between the controlled vehicle and other vehicles. Once a collision occurs, the mission will be terminated and accumulated a negative reward of $\kappa$.

\begin{table*}[t]
	\centering
	\caption{Results of experiments. Indicators include the mean, variance and final convergence value of the entire training curve. The higher the Mean and Final, the better; the lower the Std, the better.}
	\scalebox{0.84}{\begin{tabular}{lcccccccccccc}
		\toprule
		\textbf{Method} & \multicolumn{3}{c}{\textbf{Highway}} & \multicolumn{3}{c}{\textbf{SUMO}} & \multicolumn{3}{c}{\textbf{Pendulum}} & \multicolumn{3}{c}{\textbf{Lunar lander}}                                                                                                                                                          \\
		\cmidrule(lr){2-4} \cmidrule(lr){5-7} \cmidrule(lr){8-10} \cmidrule(lr){11-13}
		                & \textbf{Mean}                        & \textbf{Std}                      & \textbf{Final}                        & \textbf{Mean}                             & \textbf{Std}  & \textbf{Final}     & \textbf{Mean}      & \textbf{Std}  & \textbf{Final}     & \textbf{Mean}      & \textbf{Std}  & \textbf{Final}     \\
		\midrule

		CEA (ours)      & $\mathbf{29.6}$                      & 17.2                              & $\mathbf{51.5}$                       & \underline{-994.3}                        & 1134.7        & $\mathbf{-185.5}$  & -677.8             & 459.3         & -197.8             & \underline{-157.1} & 77.9          & $\mathbf{-94.3}$   \\
		CEA+PER (ours)  & 25.7                                 & 14.1                              & \underline{41.5}                      & -1090.3                                   & 1125.4        & \underline{-245.4} & \underline{-615.2} & 474.9         & \underline{-172.5} & -239.7             & 80.2          & -238.4             \\
		MBPO            & 4.7                                  & 2.7                               & 3.7                                   & -2455.2                                   & 719.1         & -3009.6            & -                  & -             & -                  & -                  & -             & -                  \\
		RDQN            & 24.7                                 & 14.4                              & 13.5                                  & -1209.5                                   & 1111.3        & -280.7             & -                  & -             & -                  & -                  & -             & -                  \\
		SAC             & 9.8                                  & 6.75                              & 9.6                                   & -2603.1                                   & 1554.3        & -1327.5            & -624.6             & 401.7         & -345.7             & -212.0             & 62.5          & -239.3             \\
		PPO             & \underline{29.5}                     & 9.1                               & 34.4                                  & $\mathbf{-875.7}$                         & 524.7         & -389.9             & -1151.3            & 171.3         & -992.2             & -157.3             & 89.8          & \underline{-112.7} \\
		DDPG            & -                                    & -                                 & -                                     & -                                         & -             & -                  & $\mathbf{-516.7}$  & 468.9         & $\mathbf{-146.5}$  & $\mathbf{-143.2}$  & 105.4         & -113.0             \\
		\bottomrule
	\end{tabular}}
	\label{table:expsummary}

	\centering
	\caption{Ablation experiments on PER and CEA. Backbones of Highway and SUMO is RDQN, and the backbone of the other two is DDPG.}\label{table:ablation}
	{\footnotesize
	\begin{tabular}{lcccccccc}
		\toprule
		\textbf{Method}         & \multicolumn{2}{c}{\textbf{Highway}} & \multicolumn{2}{c}{\textbf{SUMO}} & \multicolumn{2}{c}{\textbf{Pendulum}} & \multicolumn{2}{c}{\textbf{Lunar lander}}                                                                                     \\
		\cmidrule(lr){2-3} \cmidrule(lr){4-5} \cmidrule(lr){6-7} \cmidrule(lr){8-9}
		                        & \textbf{Mean}                        & \textbf{Final}                    & \textbf{Mean}                         & \textbf{Final}                            & \textbf{Mean}      & \textbf{Final}     & \textbf{Mean}      & \textbf{Final}     \\
		\midrule
		backbone+CEA (ours)     & $\mathbf{29.6}$                      & $\mathbf{51.5}$                   & $\mathbf{-994.3}$                     & $\mathbf{-185.5}$                         & -667.8             & -197.8             & \underline{-157.1} & $\mathbf{-94.3}$   \\
		backbone+CEA+PER (ours) & \underline{25.7}                     & \underline{41.5}                  & -1090.3                               & -245.4                                    & -615.2             & -172.5             & -239.7             & -238.4             \\
		backbone+PER            & 22.2                                 & 9.8                               & -1209.5                               & -280.7                                    & $\mathbf{-493.8}$  & $\mathbf{-152.9}$  & -221.9             & -191.0             \\
		backbone                & 24.7                                 & 13.5                              & \underline{-1018.4}                   & \underline{-234.2}                        & \underline{-516.7} & \underline{-146.5} & $\mathbf{-143.2}$  & \underline{-113.0} \\
		\bottomrule
	\end{tabular}}
\end{table*}

\subsection{Comparison algorithms}

The algorithms under examination in this study encompass model-free methods including RDQN, Proximal Policy Optimization (PPO), DDPG and SAC. While Model-Based Policy Optimization (MBPO) \cite{MBPO} is compared as well, which represents SOTA in model-based RL. RDQN, a potent baseline, combines multiple effective tricks with DQN. Although RDQN itself has PER, in the experiment we will test PER alone as an ablation condition, so RDQN refers to RDQN without PER by default, otherwise it will be marked as RDQN+PER. PPO, the leading model-free approach for discrete actions, is renowned for its stability. SAC, initially designed for entropy RL, is one of the most advanced off-policy methods that balances exploration and exploitation. MBPO integrates an ensemble model into SAC to simulate the environment and enhance the algorithm's planning capabilities. To simplify the presentation in the experiments, in the discrete action space environment, CEA is built within the framework of RDQN, while in the continuous action space environment, CEA is built within the framework of DDPG. It is important to note that CEA remains a model-free method, as it learns solely from state transition rules and does not perform planning.

\subsection{Parameters and equipments}

The algorithms' parameters are disclosed in the Appendix \ref{ap:parameters}. We test the algorithm on the consumer-grade graphics card RTX3060 and the high-speed computing card A100 and found no obvious difference. Finally, we implement all experiments on a RTX3060 GPU. However, if the experimental environment is more complex and the amount of data is larger, we believe that this requires better equipments.

\begin{figure}[htbp]
	\centering
	\subfigure[SUMO training. Seeds are from 42 to 45.]{
		\includegraphics[width=0.4\textwidth]{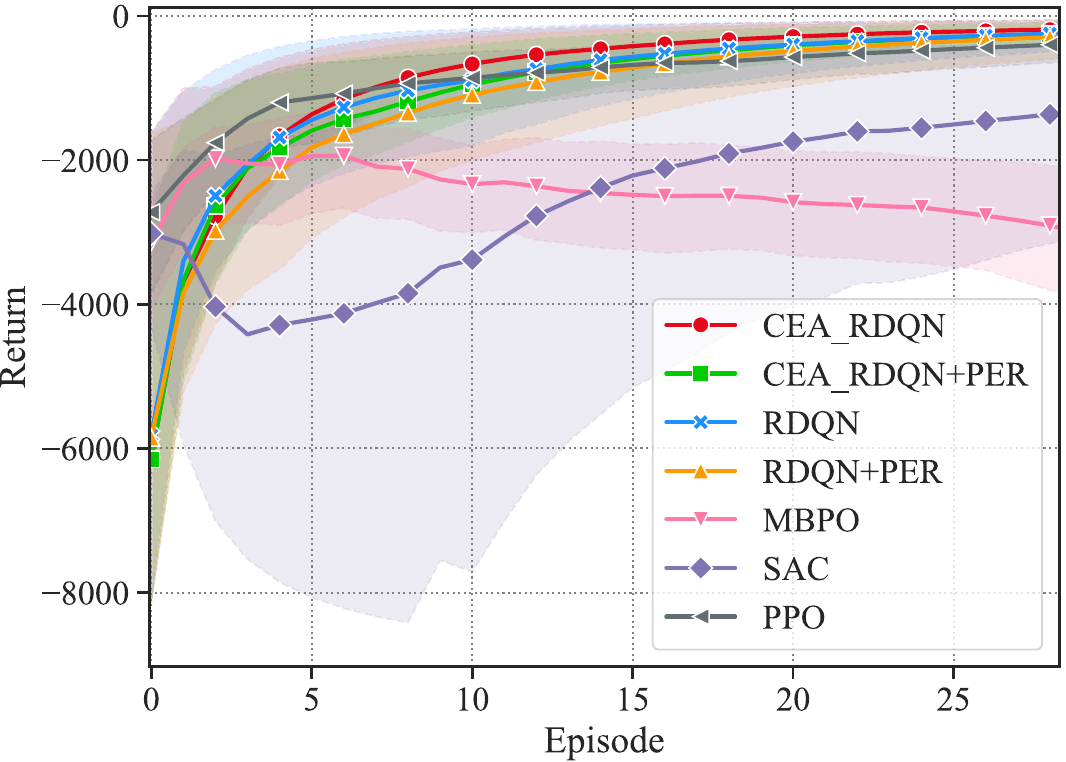}
		\label{sumo-exp}
	}
	\subfigure[Highway training. Seeds are from 42 to 45.]{
		\includegraphics[width=0.4\textwidth]{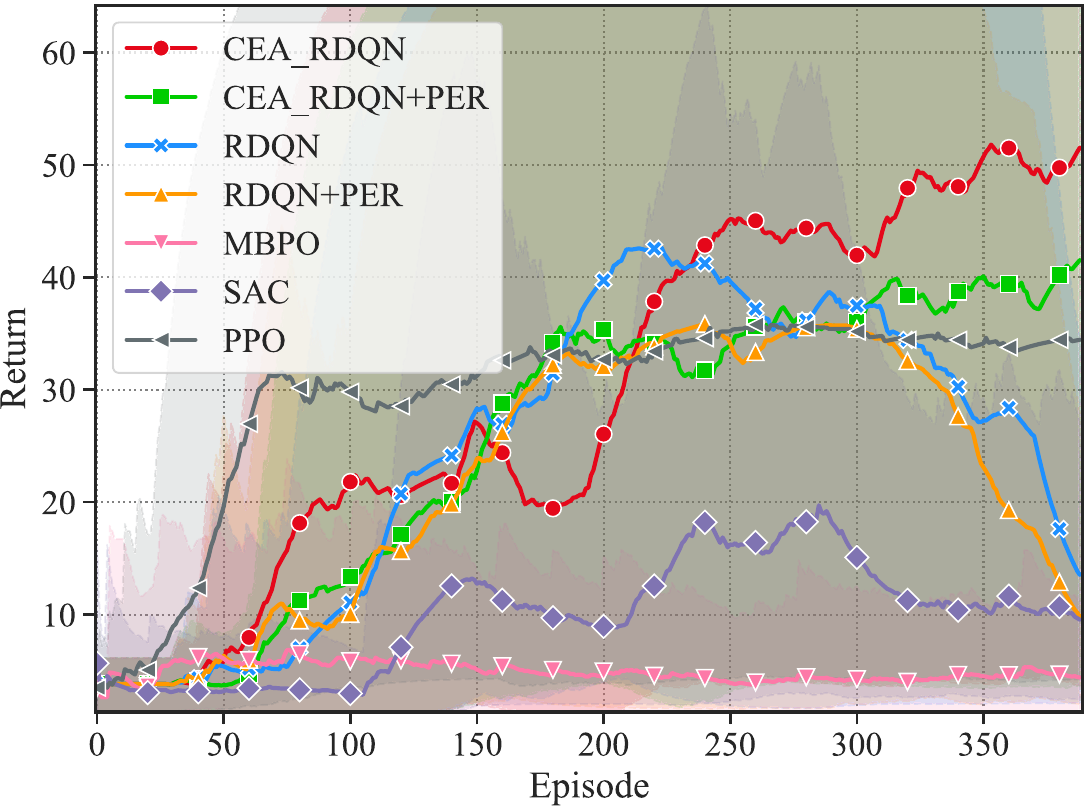}
		\label{highway-exp}
	}
	\subfigure[Pendulum training. Seeds are from 1 to 7.]{
		\includegraphics[width=0.40\textwidth]{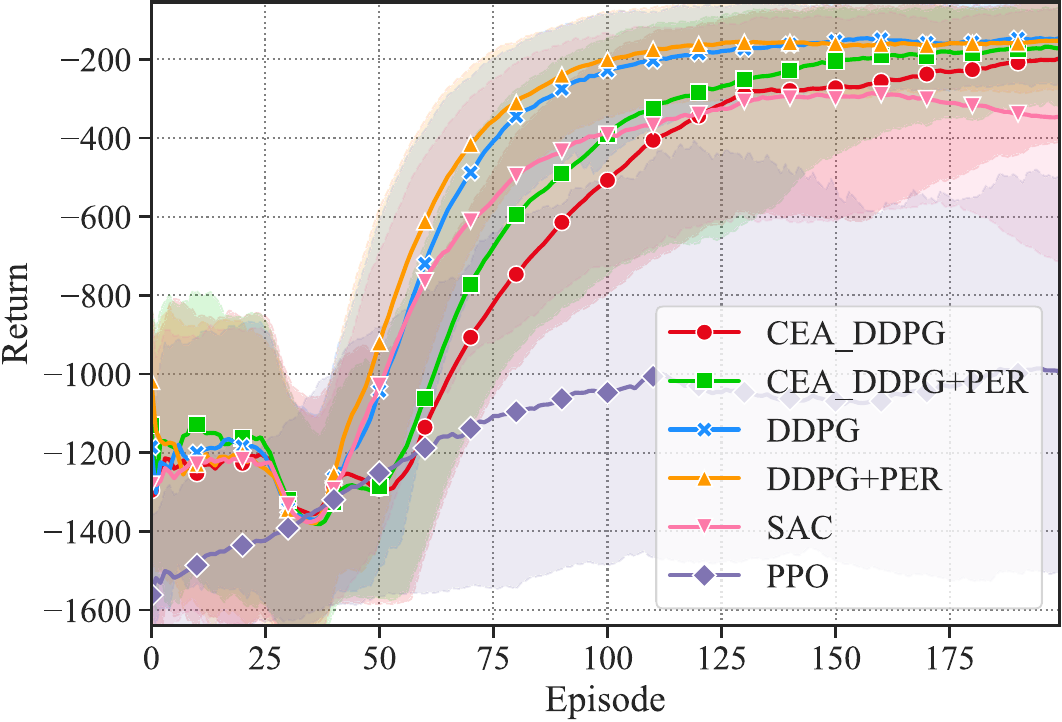}
		\label{pendulum-exp}
	}
	\subfigure[Lunar lander training. Seeds are from 1 to 7.]{
		\includegraphics[width=0.40\textwidth]{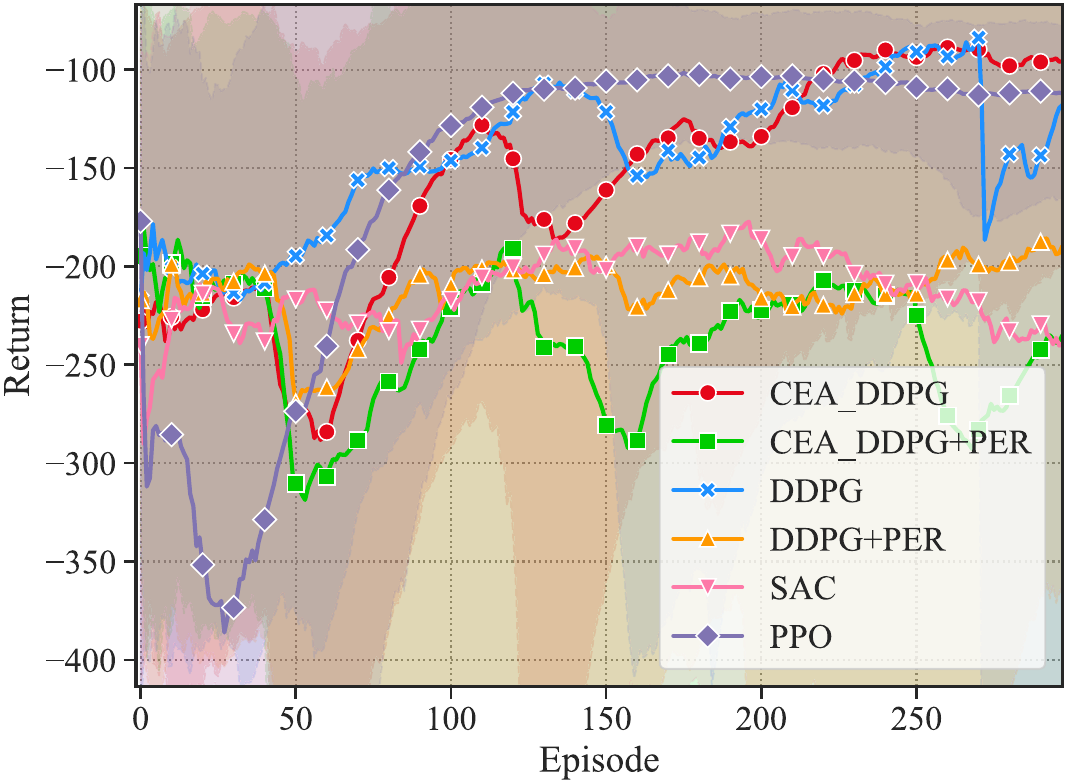}
		\label{lunar-exp}
	}
	\caption{Training process of all experimental environments} \label{all-exp}
\end{figure}

\section{Experimental results} \label{Experiment}

This section will present results of comparative and ablation experiments. To ensure fairness, all algorithms are trained using multiple seeds to minimize randomness. All seeds will be disclosed in the title of the experiment figures. Different algorithms are marked with different colors and markers. The shaded area of each algorithm represents the maximum and minimum return of each training set in different seeds, while the solid line shows the average value. All curves display exponential moving averages with a smoothing factor of 0.05.

Referring to Fig. \ref{sumo-exp} for the training performance in SUMO, the algorithm proposed in this paper is the curve CEA. In the traffic signal control mission, most algorithms converge to similar optimal levels, differing mainly in convergence speed and control stability. Notably, MBPO fails to converge, as it struggles with generalizing reward predictions, which leads to a poor model and inadequate guidance for the training of RL agent. The visualization of the model training results of MBPO are shown in Subsection \ref{failmbpo}.

Referring to Fig. \ref{highway-exp} for Highway benchmark. PPO exhibits the quickest convergence speed, but its final performance level falls short compared to CEA. Additionally, RDQN encounters a noticeable catastrophic forgetting issue in the last period training. In Highway, initial environment conditions occasionally lead to vehicles being too close, causing the algorithm to struggle in controlling promptly. This results in frequent collisions and low returns, manifesting as a large shaded area in the figure.

Refer to Fig \ref{pendulum-exp}, in the pendulum task, CEA performs poorly. The training effect of DDPG with CEA module is not as good as DDPG itself. We think it may be because the environment does not meet the assumption of mutual simulation, or it may be related to insufficient training of STA.

The experimental results of the Lunar lander task refer to Fig. \ref{lunar-exp}, CEA still maintains a good performance, indicating that it can still play a role in the continuous action space. Combining the results of pendulum and lunar lander, we believe that we need to further consider how to define the soft bisimulation assumption and consider further improvements to CEA.

We present experimental result in Table \ref{table:expsummary}. The boldface indicates the best performance, and the underlined indicates the second-best performance. It can be observed that CEA has an excellent overall performance. The second is DDPG, but DDPG cannot support discrete action space tasks well.

The ablation experiments are shown in Table \ref{table:ablation}, which mainly focus on backbone, CEA and PER. It is observed that PER does not necessarily play a critical role, but CEA maintains the performance improvement. We believe that when the reward signal is dense, PER may not lead to improvement.

\begin{wrapfigure}[20cm]{r}{0.5\linewidth}
	\centering
	\includegraphics[width=\linewidth]{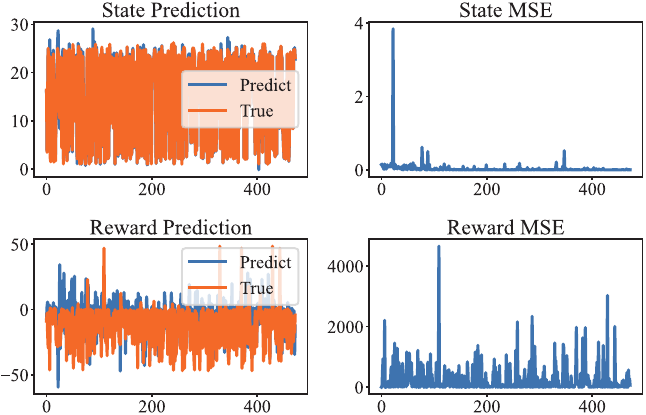}
	\caption{The training of MBPO's environment model on the SUMO. MBPO is committed to maintaining a virtual environment composed of an integrated neural network model, which accepts the input of the current state and action, and then outputs the prediction and reward value of the next state. It is observed from the figure that the model learns the state transfer law well, but it is difficult to accurately predict the reward value.}\label{MBPO-model}
\end{wrapfigure}

The following section is about innovative study, focusing on the details of the experiments and innovative methods in this paper.

\section{Innovation study} \label{Innovation}

In this section, we will critically examine the limitations of MBPO and delve into the specifics of our novel contributions, which consist of the STA and CEA methodologies.

\subsection{The deficiency of model-based RL}\label{failmbpo}

Here we denote the virtual environmental model used in model-based RL as model. The primary challenge lies in maintaining model accuracy in dynamic environments, where frequent updates are necessary but can compromise prediction precision. Our observations reveal that while models can learn state transitions relatively well in non-stationary settings like traffic but struggle to capture reward signals, which can significantly impair model training. This issue is illustrated in Fig. \ref{MBPO-model}. Consequently, this motivates the development of STA and CEA.

\begin{wrapfigure}[46cm]{r}{0.45\textwidth}
	\centering
	\includegraphics[width=0.9\linewidth]{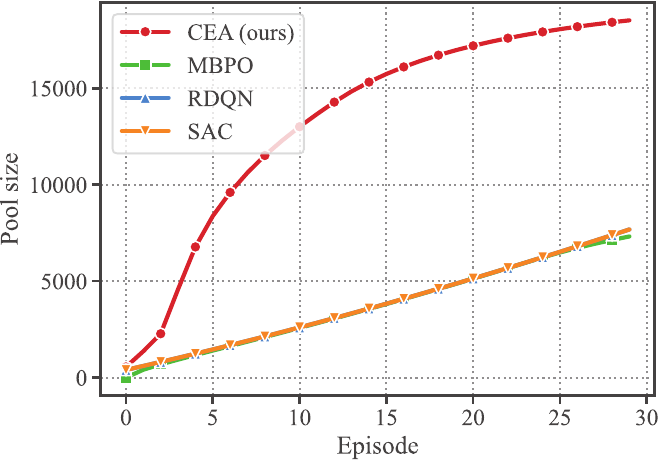}
	\caption{ Experience pool size in SUMO mission. It can be observed that through the augmentation of counterfactual experience, the volume of the experience pool rapidly expands, but the rate gradually decreases, and the rate is adjustable.}\label{sumo-pool-size}
    \vspace{0.5cm}
	\centering
	\includegraphics[width=0.9\linewidth]{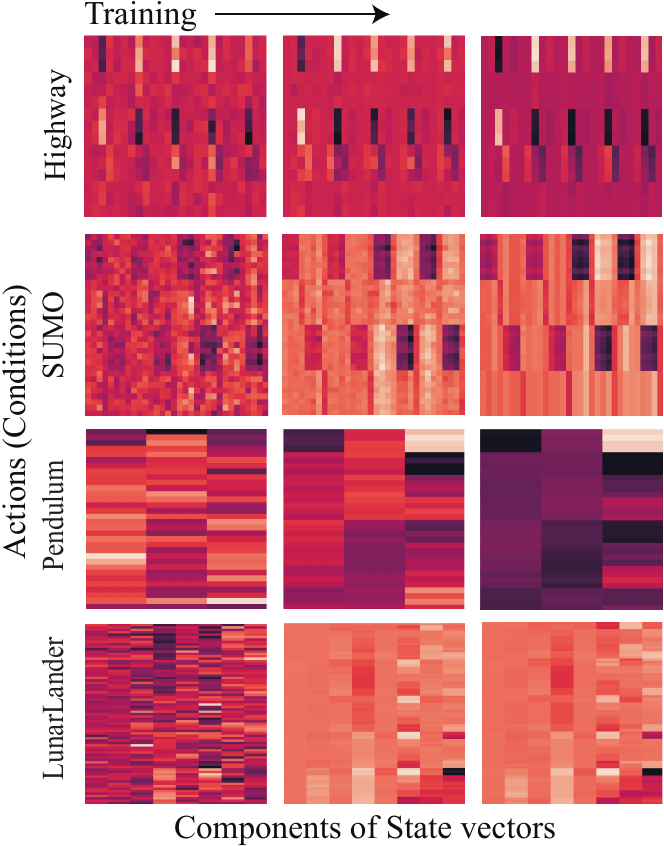}
	\caption{STA generative metrics. From left to right represents the training period. On the vertical axis, we give various actions and put the same actions together. While the transfer vectors generated by the same actions are similar, which should show color blocks. It can be observed that the color blocks are not obvious at the beginning of training, but become more obvious as training progresses, indicating that STA has learned the state transition pattern.}\label{sta-generation}
\end{wrapfigure}

\subsection{Generation of state transition autoencoder}

The STA generates state transition vectors based on counterfactual actions. Fig. \ref{sta-generation} shows generative matrices for different tasks: initial training on the left and final results on the right, with three representative matrices from each stage. The vertical axis represents counterfactual actions, grouped and ordered, while the horizontal axis shows components of state transition vectors. Each row of colored blocks is a complete vector. As training progresses, the autoencoder better captures state transition dynamics, shown as clusters of dark or light blocks. The randomness of vectors reflects non-stationarity of environments, justifying the use of variational autoencoders.

For experience augmentation as shown in Fig. \ref{sumo-pool-size}, CEA expands experiences rapidly initially but slows as experience accumulates. Once a threshold is reached, we limit adding counterfactual experience, focusing on real experiences. When the experience pool is full, counterfactual experiences are no longer added. SAC and MBPO have identical experience growth rates, leading to almost overlapping curves in the figure.

\begin{figure}[t]
	\centering
	\subfigure[Kernel density estimation of experience features in SUMO. Virtual exp represent counterfactual experience.]{
		\includegraphics[width=0.45\textwidth]{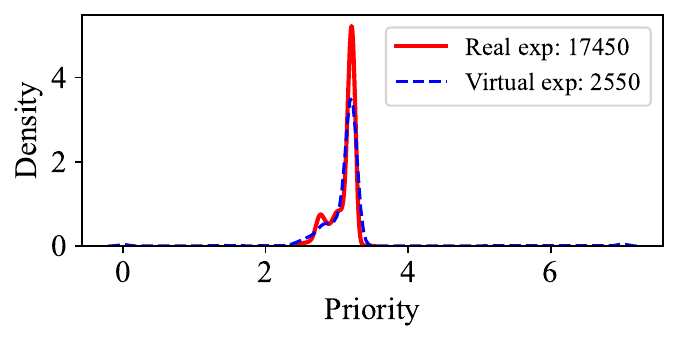}
		\label{sumo-staexp}
	}
	\subfigure[Kernel density estimation of experience features in Highway. Virtual exp represent counterfactual experience.]{
		\includegraphics[width=0.45\textwidth]{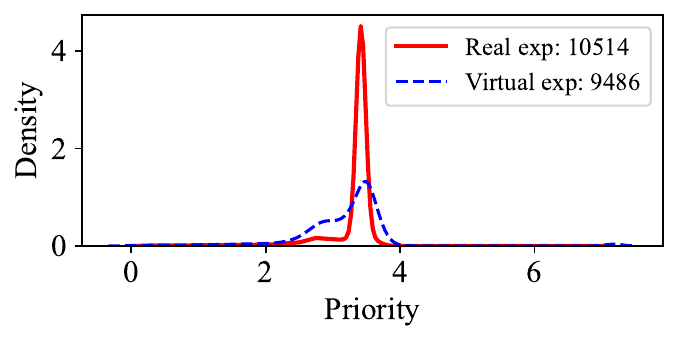}
		\label{highway-staexp}
	}
	\caption{Kernel density estimation. The legend gives quantity of the real experience and the counterfactual experience. The horizontal axis represents the priority of PER calculation. The probability density distribution of real experiences and counterfactual experiences, i.e., virtual experiences, can be observed to be aligned in the figure, indicating that the generation quality of counterfactual experiences is relatively high.} \label{sumo-highway-staexp}
\end{figure}

Finally, although PER is not always helpful for CEA, it still reveals the similarity of priority distribution between counterfactual experience and real experience, refer to Fig. \ref{sumo-highway-staexp}. We show the kernel density distribution of priorities for two different experiences in SUMO and Highway. The two experiences have similar distribution shapes, indicating the rationality of the construction of counterfactual experience.

\section{Discussion and future work} \label{discussion}

Although our method performs well in many scenarios, there are several challenges with the proposed approach.

The complexity of CEA is a significant concern, since we need to perform maximum entropy optimization sampling via Gaussian kernel density estimation. High-dimensional Gaussian kernel density estimation is usually time-consuming. As well as the computational burden of the CTP, which involves comparing large volumes of counterfactual experience with real data. These challenges are especially pronounced in large experience pools, where time efficiency can be compromised.

In addition, it may also be necessary to consider the rate adjustment of introducing counterfactual experiences, because too few real experiences are difficult to train high-quality STA models for prediction, and when real experiences reach a certain amount, the introduction of counterfactual experiences should be reduced. As shown in Fig. \ref{sumo-pool-size}, we adopted annealing-style experience supplementation, where the amount of supplementation gradually decreases, which can be further designed into an adaptive form.

\section{Conclusion} \label{conclusion}

In this study, we present a comprehensive CEA algorithm, which employs maximum entropy optimization in kernel density estimation to sample counterfactual actions. These actions are then used as generative conditions in the STA to predict subsequent states. STA can be trained effectively using only data from random interactions with the environment, making it easy to train. CEA then uses distance matching to extract suitable reward signals from real-world experiences, resulting in complete counterfactual experiences and storing in the experience pool. We evaluated CEA across various scenarios, where it demonstrated superior overall performance. Our research explores both the practical application and the innovative aspects of STA and CEA, offering valuable insights for future work.

\section*{Acknowledgment}
This work was supported by Hubei Provincial Key Laboratory of Metallurgical Industry Process System Science (Y202105). This work is also supported by High-Performance Computing Center of Wuhan University of Science and Technology.

\section*{Declaration}

\begin{itemize}
	\item \textbf{Conflict of Interest} We declare that we have no known competing financial interests or personal relationships that could have appeared to influence the work reported in this paper.
	\item \textbf{Data Availability} The datasets generated during and/or analyzed during the current study are available from the corresponding author on reasonable request.
	\item \textbf{Ethical approval} Our study did not raise any ethical questions, i.e., none of the subjects were humans or living individuals.
	\item \textbf{Informed consent} All authors are aware of this article and agree to its submission.
	\item \textbf{Declaration of generative AI and AI-assisted technologies in the writing process} During the preparation of this work, the first author used generative AI in order to proofread grammar and expression. After using this service, the first author reviewed and edited the content as needed and take full responsibility for the content of the publication.
\end{itemize}


\bibliographystyle{unsrt}

\bibliography{cas-refs}



\appendix 
\setcounter{table}{0} 
\setcounter{equation}{0} 
\section{Calculation of evidence lower bound}\label{ap:proof1}

Here we provide the calculation process of $\mathbb{D}_{KL}\left[q(z|d,a)||p(z|a)\right]$, where $q(z|d,a)$ follows $N(\mu, \sigma)$, where $\mu$ and $\sigma$ depend on the output of encoder. If we input multiple conditions for batch calculation, it should be a multivariate normal distribution, but we can give the calculation for a univariate normal distribution to generalize.

Here we substitute the normal distribution probability density function in $\mathbb{D}_{KL}\left[q(z|d,a)||p(z|a)\right]$:
\begin{equation}
	\begin{aligned}
	& \mathbb{D}_{KL}{\left(N(\mu,\sigma^2){\left\|N(0,1)\right.}\right)} \\
	& =\int\frac{1}{\sqrt{2\pi\sigma^2}}e^{-(x-\mu)^2/2\sigma^2}\left(\log\frac{e^{-(x-\mu)^2/2\sigma^2}/\sqrt{2\pi\sigma^2}}{e^{-x^2/2}/\sqrt{2\pi}}\right)dx \\
	& =\int\frac{1}{\sqrt{2\pi\sigma^2}}e^{-(x-\mu)^2/2\sigma^2}\log\left\{\frac{1}{\sqrt{\sigma^2}}\mathrm{exp}{\left\{\frac{1}{2}\left[x^2-(x-\mu)^2/\sigma^2\right]\right\}}\right\}dx \\
	& =\frac{1}{2}\int\frac{1}{\sqrt{2\pi\sigma^2}}e^{-(x-\mu)^2/2\sigma^2}\left[-\log\sigma^2+x^2-(x-\mu)^2/\sigma^2\right]dx.
   \end{aligned}
\end{equation}

We take apart the square brackets on the right side of the integral and calculate them separately. The first is:
\begin{equation}
	\begin{aligned}
	&\int\frac{1}{\sqrt{2\pi\sigma^2}}e^{-(x-\mu)^2/2\sigma^2}\left[-\log\sigma^2\right]dx\\
	&=\left[-\log\sigma^2\right]\int\frac{1}{\sqrt{2\pi\sigma^2}}e^{-(x-\mu)^2/2\sigma^2}dx\\
	&=-\log\sigma^2.
	\end{aligned}
\end{equation}

The second is:
\begin{equation}
	\begin{aligned}
		&\int\frac{x^2}{\sqrt{2\pi\sigma^2}}e^{-(x-\mu)^2/2\sigma^2} dx\\
		&=\mathbb{E}_{x\sim N(\mu, \sigma^2)}(x^2)\\
		&=\mu^2+\sigma^2.
	\end{aligned}
\end{equation}

The third is:
\begin{equation}
	\begin{aligned}
		&\int\frac{1}{\sqrt{2\pi\sigma^2}}e^{-(x-\mu)^2/2\sigma^2}[-(x-\mu)^2/\sigma^2] dx\\
		&=-\mathbb{E}_{x\sim N(\mu, \sigma^2)}(x-\mu^2)/\sigma^2\\
		&=-\sigma^2/\sigma^2\\
		&=-1.
	\end{aligned}
\end{equation}

Therefore we give
\begin{equation}
	\begin{aligned}
		\mathbb{D}_{KL}\left[q(z|d,a)||p(z|a)\right] = \frac{1}{2}\sum(-\log{\sigma^2}+\mu^2+\sigma^2-1).
	\end{aligned}
\end{equation}

\section{Parameters of all algorithms}\label{ap:parameters}
Here we provide the parameter settings for all algorithms in the experiment, as shown in Fig. \ref{tab:combined_parameters}. It should be noted that in SAC-discrete parameter settings, target\_entropy is calculated by $0.98(-\log(1 / action\_space))$. The same is true in MBPO. And the RL backbone of MBPO is SAC-discrete.
\begin{table}
	\centering
	\caption{Parameter settings for Rainbow DQN, SAC-discrete, CEA, PPO, DDPG and MBPO algorithms respectively.}
	\label{tab:combined_parameters}
	\scalebox{0.8}{\begin{tabular}{llll}
		\toprule
		\textbf{Model}      & \textbf{Parameter} & \textbf{Value} & \textbf{Description}                  \\
		\midrule
		Rainbow DQN         & gamma              & 0.99           & Discount factor for future rewards     \\
		                     & alpha              & 0.2            & Determines how much prioritization is used \\
		                     & beta               & 0.6            & Determines how much importance sampling is used \\
		                     & prior\_eps         & 1e-6           & Guarantees every transition can be sampled \\
		                     & v\_min             & 0              & Min value of support                   \\
		                     & v\_max             & 200            & Max value of support                   \\
		                     & atom\_size         & 51             & The unit number of support             \\
		                     & memory\_size       & 20000          & Size of the replay buffer              \\
		                     & batch\_size        & 128            & Batch size for updates                 \\
		                     & target\_update     & 100            & Period for target model's hard update  \\
		\midrule
		SAC-discrete        & actor\_lr          & 5e-4           & Learning rate for the actor network    \\
		                     & critic\_lr         & 5e-3           & Learning rate for the critic network   \\
		                     & alpha\_lr          & 1e-3           & Learning rate for the temperature parameter \\
		                     & hidden\_dim        & 128            & Dimension of hidden layers             \\
		                     & gamma              & 0.98           & Discount factor for future rewards     \\
		                     & tau                & 0.005          & Soft update parameter                  \\
		                     & buffer\_size       & 20000          & Size of the replay buffer              \\
		                     & target\_entropy    & 1.36           & Target entropy for the policy          \\
		                     & model\_alpha       & 0.01           & Weighting factor in the model loss function \\
		                     & total\_epochs      & 1              & Total number of training epochs        \\
		                     & minimal\_size      & 500            & Minimum size of the replay buffer before updating \\
		                     & batch\_size        & 64             & Batch size for updates                 \\
		\midrule
		CEA                 & memory\_size       & 20000          & Size of the replay buffer              \\
		                     & batch\_size        & 128            & Batch size for updates                 \\
		                     & target\_update     & 100            & Period for target model's hard update  \\
		                     & threshold\_ratio   & 0.1            & Threshold ratio for choosing CTP       \\
		\midrule
		PPO                 & actor\_lr            & 3e-4            & Learning rate for the actor 	network   \\
		                     & critic\_lr           & 3e-4           & Learning rate for the critic network  \\
		                     & gamma                & 0.99           & Discount factor for future rewards    \\
		                     & total\_epochs        & 1              & Number of training iterations         \\
		                     & total\_episodes      & 100            & Number of games played per training iteration \\
		                     & eps                  & 0.2            & Clipping range parameter for the PPO objective (1 - eps to 1 + eps) \\
		                     & epochs               & 10             & Number of epochs per training sequence in PPO \\
		\midrule
		MBPO                & real\_ratio          & 0.5            & Ratio of real to model-generated data \\
		                     & actor\_lr            & 5e-4           & Learning rate for the actor network   \\
		                     & critic\_lr           & 5e-3           & Learning rate for the critic network  \\
		                     & alpha\_lr            & 1e-3           & Learning rate for the temperature parameter \\
		                     & hidden\_dim          & 128            & Dimension of hidden layers            \\
		                     & gamma                & 0.98           & Discount factor for future rewards    \\
		                     & tau                  & 0.005          & Soft update parameter                 \\
		                     & buffer\_size         & 20000          & Size of the replay buffer             \\
		                     & target\_entropy      & 1.36           & Target entropy for the policy         \\
		                     & model\_alpha         & 0.01           & Weighting factor in the model loss function \\
		                     & rollout\_batch\_size & 1000           & Batch size for rollouts               \\
		                     & rollout\_length      & 1              & Length of the model rollouts          \\
		\midrule
		DDPG			 &	actor\_lr          & 2e-4    & Learning rate of actor network \\
						 &	critic\_lr         & 2e-4    & Learning rate of critic network \\
						 &	hidden\_dim        & 128                 & Dimension of hidden layers \\
						 &	buffer\_size       & 5e4       & Size of replay buffer \\
						 &	minimal\_size      & 5e3       & Minimum sample size \\
						 &	gamma              & 0.98                & Discount factor \\
						 &	sigma              & 0.01                & Standard deviation of Gaussian noise \\
						 &	tau                & 0.005               & Soft update parameter \\
						 &  batch\_size        & 128                 & Batch size for training \\
		\bottomrule
	\end{tabular}}

\end{table}

\end{document}